\documentclass[sigconf]{acmart}

\AtBeginDocument{%
  \providecommand\BibTeX{{%
    \normalfont B\kern-0.5em{\scshape i\kern-0.25em b}\kern-0.8em\TeX}}}

\setcopyright{acmcopyright}
\copyrightyear{2019}
\acmYear{2019}
\acmDOI{}

\acmConference[Anchorage '19]{Anchorage '19: KDD Workshop on Applied Data Science for Healthcare}{August 5, 2019}{Anchorage, AK, USA}
\acmBooktitle{Anchorage '19: KDD Workshop on Applied Data Science for Healthcare,
  August 5, 2019, Anchorage, AK, USA}
\acmPrice{}
\acmISBN{}

\begin{document}

\title[]{Predicting assisted ventilation in Amyotrophic Lateral Sclerosis using a mixture of experts and conformal predictors}

\author{Telma Pereira}
\affiliation{%
  \institution{LASIGE, Departamento de Inform\'atica, Faculdade de ci\^encias, Universidade de Lisboa and Instituto Superior T\'ecnico, Universidade de Lisboa}
  \city{Lisbon}
  \country{Portugal}
}
  \email{telma.pereira@tecnico.ulisboa.pt }
\author{Sofia Pires}
\affiliation{%
  \institution{LASIGE, Departamento de Inform\'atica, Faculdade de ci\^encias, Universidade de Lisboa and Instituto de Medicina Molecular, Instituto de Fisiologia, Faculdade de Medicina, Universidade de Lisboa}
  \city{Lisbon}
  \country{Portugal}}
\email{sofiaferropires@gmail.com}

\author{Marta Gromicho}
\affiliation{%
  \institution{Instituto de Medicina Molecular, Instituto de Fisiologia, Faculdade de Medicina, Universidade de Lisboa}
  \city{Lisbon}
  \country{Portugal}
}
\email{martalgms@gmail.com}

\author{Susana Pinto}
\affiliation{%
  \institution{Instituto de Medicina Molecular, Instituto de Fisiologia, Faculdade de Medicina, Universidade de Lisboa}
  \city{Lisbon}
  \country{Portugal}
}
\email{susana.c.pinto@medicina.ulisboa.pt}

\author{Mamede de Carvalho}
\affiliation{%
  \institution{Instituto de Medicina Molecular, Instituto de Fisiologia, Faculdade de Medicina, Universidade de Lisboa and Department of Neurosciences and Mental Health, Hospital de Santa Maria, CHLN}
  \city{Lisbon}
  \country{Portugal}
}
\email{mamedealves@medicina.ulisboa.pt}

\author{Sara C. Madeira}
\affiliation{%
  \institution{LASIGE, Departamento de Inform\'atica, Faculdade de ci\^encias, Universidade de Lisboa}
  \city{Lisbon}
  \country{Portugal}}
\email{sacmadeira@ciencias.ulisboa.pt}

\renewcommand{\shortauthors}{Pereira et al.}

\begin{abstract}
Amyotrophic Lateral Sclerosis (ALS) is a neurodegenerative disease characterized by a rapid motor decline, leading to respiratory failure and subsequently to death. In this context, researchers have sought for models to automatically predict disease progression to assisted ventilation in ALS patients. However, the clinical translation of such models is limited by the lack of insight 1) on the risk of error for predictions at patient-level, and 2) on the most adequate time to administer the non-invasive ventilation. To address these issues, we combine Conformal Prediction (a machine learning framework that complements predictions with confidence measures) and a mixture experts into a prognostic model which not only predicts whether an ALS patient will suffer from respiratory insufficiency but also the most likely time window of occurrence, at a given reliability level. Promising results were obtained, with near 80\% of predictions being correctly identified.
\end{abstract}


\maketitle
\vspace{-1mm}
\section{Introduction}
Amyotrophic Lateral Sclerosis (ALS) is a neurodegenerative disease, commonly characterized by loss of movement due to the degeneration of motor neurons in the brain and spinal cord \cite{Brown2017}. Its severe nature and fast progression result in short survival times, averaging 3 to 5 years \cite{Brown2017}. Death in ALS patients is usually associated with respiratory failure \cite{Lechtzin2018}. Therefore, treatments designed to manage respiratory system related symptoms, such as the administration of Non-Invasive Ventilation (NIV), have shown to improve prognosis and extend survival time \cite{Bourke2006}.

In this context, an early prediction of patients' need for assisted ventilation would have significant implications for patients' quality of life and health costs. Machine Learning-based approaches have been followed to tackle this problem \cite{Carreiro2015, Pires2018}. Carreiro et al. \cite{Carreiro2015} developed a supervised learning approach to predict the need for NIV in ALS patients within specific time windows. Furthermore, Pires et al. \cite{Pires2018} attempted to improve these predictive models by tackling the heterogeneity in ALS patients, using stratified disease progression groups.

Despite the promising results of these prognostic models, their translation into clinical practice is hampered by the lack of insight on the risk of error for predictions at instance-level (in this case, at patient-level). In other words, for prognostic models to become actionable in the clinicians' decision-making process, they must provide not only the most likely prediction for a given patient but also an indication of how reliable that prediction is. 

Conformal Prediction (CP) is a machine learning framework built on top of standard classifiers that, for a given test instance, computes a p-value for each possible class, which can be used as a confidence measure (indicating the likelihood of that prediction being correct), or to produce a prediction set guaranteed to include the true label, at that confidence level. CP has been successfully applied in health-related domains \cite{Papadopoulos2009, Lambrou2010}.

In this work, we evaluate the feasibility of the CP framework to target the reliability of predictions at patient-level when predicting NIV for ALS patients. In particular, we propose a prognostic model using a mixture of experts (i.e., models learned with different time windows) which not only predicts whether a given patient will suffer from respiratory insufficiency but also outputs the most likely time window of occurrence, at a given reliability level. To the best of our knowledge, there are no previous studies on CP on ALS prediction, making it a relevant case study. Furthermore, we use a different way of building learning examples using time windows when compared with previous literature \cite{Carreiro2015, Pires2018}, by narrowing the time width of prediction. 
 
\vspace{-1mm}
\section{Dataset and Methods}

\subsection{Data}
\label{datasect}
We used a cohort of 1360 ALS patients, followed in the ALS clinic of the Translational Clinical Physiology Unit, Hospital de Santa Maria, Lisbon, from 1992 until March 2019. It comprises 27 variables, comprising demographic, clinical (including respiratory tests and neurophysiological data), and genetic data. This data is collected from every participant at the baseline assessment, as well as on their quarterly follow-up consultations. For a detailed description of these variables we refer to \cite{Pires2018}.
\vspace{-1.5mm}
\subsubsection{Creating learning examples}
\label{createLE}

Since data may not all be collected in the same day, a preprocessing step is required to merge all features into a single observation, called here as a snapshot, reflecting the summary of the patient condition around that appointment's time. We followed the approach proposed in \cite{Carreiro2015}, a bottom-up hierarchical clustering with constraints strategy to cluster temporally-related tests. In the end, we can have multiple snapshots per patient regarding different appointments. 

Then, we stratified the created patients' snapshots regarding their time of progression to respiratory insufficiency. These correspond to the learning examples used to train the prognostic models. This learning approach using time windows was already used in \cite{Carreiro2015, Pires2018}, and allows to answer the question of "Will an ALS patient require NIV \textit{k} days after the medical assessment?". However, these time windows are inclusive, in the sense that prognostic models built for a 365-days time window, for instance, might include cases requiring NIV either at 90 or 180 days after the assessment. For clinicians, it would be more informative if we could narrow the time window of prediction, for instance, to the temporal distance between appointments (in this case, three-months intervals), thus moving to the question "Will an ALS patient require NIV within \textit{k} and \textit{k+90} days after the medical assessment?". In previous work \cite{Carreiro2015, Pires2018}, time windows of 90, 180, and 365 days (3, 6, and 12 months respectively) were used, as recommended by clinicians. In this context, we predict the need for NIV at up to 90 days, from 90 days to 180 days, and from 180 days to 365 days.

\subsection{Cross Conformal Prediction}
Let us assume that we are given a training set ${(x_{1}, y_{1}),...,(x_{n-1}, y_{n-1})}$, where $x_{i} \in X$ is a vector of attributes and $y_{i} \in Y$ is the class label (binary classification problem). Given a new test example $(x_{n})$ we aim to predict its class and assess the level of uncertainty of such prediction. Intuitively, we assign each class $y_{n} \in Y$ to $x_{n}$, at a time, and then evaluate how (dis)similar the example $(x_{n}, y_{n})$ is in comparison with the training data, using a (non-)conformity measure. This (non-)conformity measure is a function that assesses the (dis-)similarity between examples by means of a numerical (non-)conformity score ($\alpha_{n}$), and is generally based on the underlying classifier.  To evaluate how different $x_{n}$ is from the training set we compare its non-conformity score with those of the remaining training examples $x_{j}, j=0, ..., n-1$, using the \textit{p}-value function: 

\begin{equation}
p(\alpha_{n})=\dfrac{\vert \{ j=1,...,n: \alpha_{j} \geq \alpha_{n} \} \vert }{n},
\end{equation}
\label{eq:eq1}
\vspace{-1mm}
\noindent
where $\alpha_{n}$ is the non-conformity score of $x_{n}$, assuming it is assigned to the class label $y_{n}$. If the \textit{p}-value is small, then the test example $(x_{n},y_{n})$ is non-conforming, since few examples $(x_{i},y_{i})$ had a higher non-conformity score when compared with $\alpha_{n}$. If the \textit{p}-value is large, $x_{n}$ is very conforming, since most examples $(x_{i},y_{i})$ had a higher non-conformity score when compared with $\alpha_{n}$. CP is valid under the randomness ($i.i.d$) assumption \cite{Vovk2005}. Once \textit{p}-values are computed, CP can be used in one of the following ways: \textit{i) Using forced predictions (FP):} predicts the class with the highest p-value and output its credibility (the largest p-value) and confidence (complement to 1 of the second highest p-value), or \textit{ii) Using prediction regions (PR):} For a given confidence level ($1-\epsilon$), outputs the prediction region - $T^{\epsilon}$: set of all classes with $p(\alpha_{n})> \epsilon$.

For each test example, and each possible class label, the classifier is rerun using all training examples along with the new test example, constraining its applicability on large datasets. To overcome this computational inefficiency problem, an inductive version of this framework has emerged. In this case, the training set ${(x_{1}, y_{1}),...,(x_{n-1}, y_{n-1})}$ is divided into the proper training set training set ${(x_{1}, y_{1}),...,(x_{m}, y_{m})}$ and the calibration set ${(x_{m+1}, y_{m+1}),...,(x_{n-1}, y_{n-1})}$, where $m<n-1$. The proper training set is used to train the underlying classifier, whereas the p-values are computed using only examples in the calibration set. Later on, a new approach named Cross Conformal Prediction (CCP) \cite{Vovk2015} was proposed to cope with the loss of informational efficiency of inductive CP. It consists in splitting the training set into \textit{k} folds, where one of the \textit{k} folds is used as a calibration set while the remaining \textit{k-1} folds are merged to form the proper training data.

\subsection{Prognostic model using a mixture of experts}
\begin{figure}
    \centering
    \includegraphics[width=46mm]{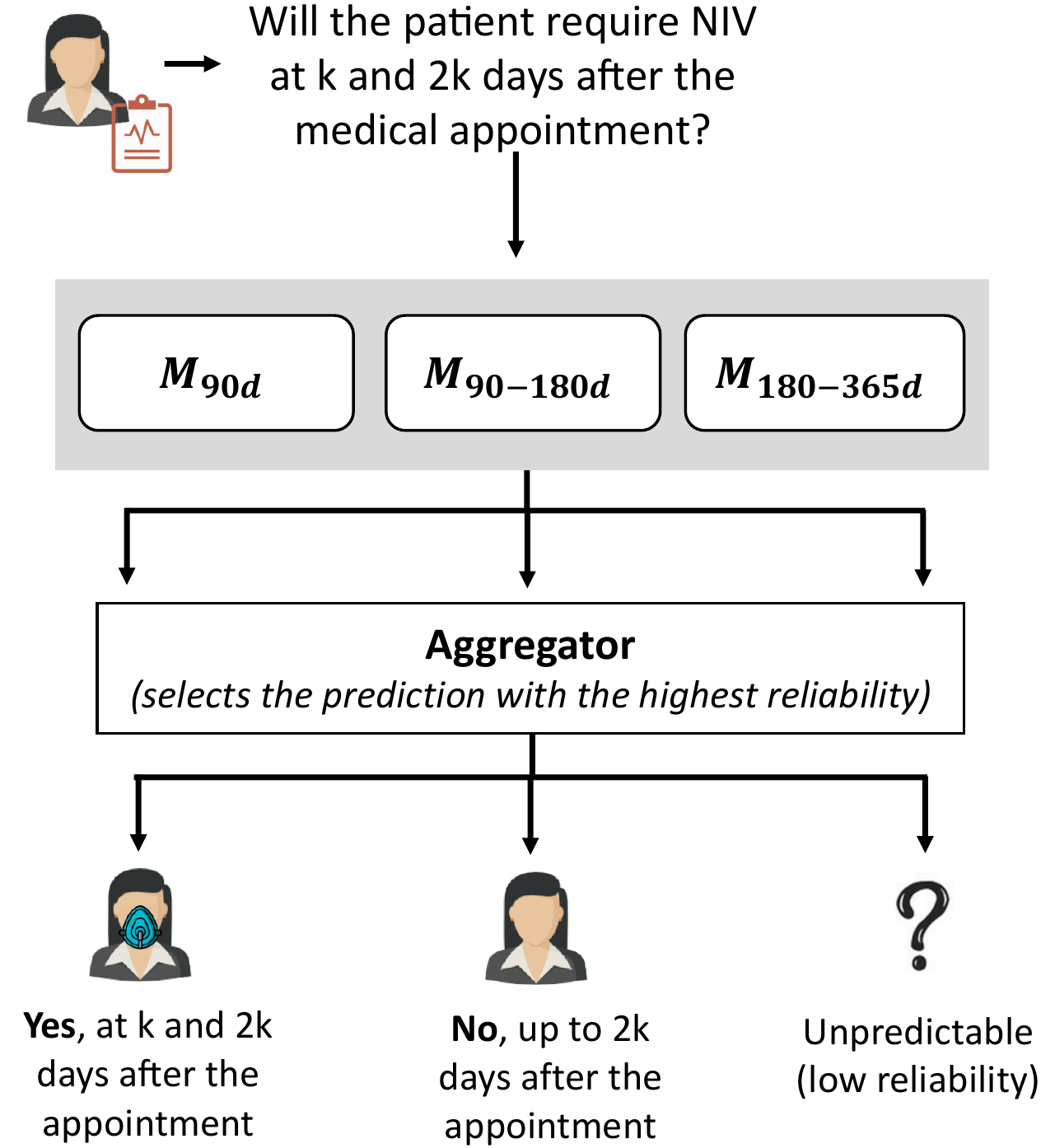}
    \caption{Workflow of the proposed prognostic model using mixture of experts to predict NIV in ALS patients.}
    \label{fig:workflow}
\end{figure}

Figure \ref{fig:workflow} depicts the supervised learning approach proposed in this work: a prognostic model in ALS using a mixture of experts. Given a new ALS patient, we predict whether he (or she) will progress to respiratory insufficiency in one of three time windows ($M_{90d}$: up to 90 days, $M_{90-180d}$: from 90 to 180 days, or $M_{180-365d}$: from 180 to 365 days) or remain stable up to the limit of each time window (these are the \textit{base models}). Therefore, we end up with three predictions and the respective reliability measure. In the next step, one must define an aggregation rule to combine those predictions into a final one. In this work, we decided to predict the class with the highest reliability (in this case, measured by the CP credibility). As such, the proposed prognostic model not only predicts whether a given ALS patient will require NIV, but also when is it more likely to happen. Moreover, a measure reflecting the reliability of the predicted class is outputted. If all predictions are below the predefined reliability threshold, we consider that case as unpredictable (No prediction). In a broader sense, we can say our prognostic model addresses a multi-class problem.

\vspace{-2.5mm}

\subsection{Classification settings}

In the evaluation of the proposed prognostic model, an \textit{outer 5-}fold cross-validation (CV) was performed on top of the \textit{inner 5} folds of the CCP. The folds were picked randomly, maintaining class proportions. For each \textit{outer} fold, we created an overall validation set by merging the testing sets from each base model (regarding different time windows). 

We tested three reliability thresholds ($\tau$=0.80, 0.90, 0.95) considered useful in clinical practice. To evaluate the classification performance of the base models $M_{90d}$, $M_{90-180d}$, and $M_{180-365d}$, we assessed the Area under the ROC curve (AUC), sensitivity, and specificity, using the testing sets created per time window (from the \textit{outer} fold). Then, to evaluate the prognostic model using a mixture of experts, we output the number of correct and incorrect predictions made, using the validation set.

We tested different classifiers within the CCP framework, using the credibility to reflect the uncertainty of predictions (Forced Predictions), and the standard classifiers alone, without predictions uncertainty, for the sake of comparability. Namely, we tested the following classifiers: Na\"ive Bayes, Support Vectors Machines with the polynomial (SVM Poly) and RBF kernel (SVM RBF), Logistic Regression, and Decision Tree with J48 algorithm.

Class imbalance was tackled by Random Undersampling and SMOTE as suggested in \cite{Pires2018}. Particularly, we first randomly undersample the majority class until a balance of 60/40\% is achieved, and then, SMOTE was used to reach a 50/50\% class proportion. 

Moreover, the most relevant set of features was selected by the feature selection ensemble algorithm proposed in \cite{Pereira2018}. This feature selection approach starts by ranking features according to their relevance as assessed by a consensus of different feature selection algorithms and then select the top-ranked features which maximize both predictability and stability performances.

The proposed prognostic model was implemented in Java using WEKA's functionalities (version 3.8.0).

\section{Results and Discussion}

The data used in this work is described in Section \ref{datasect} and summarized in Table \ref{tab:freq}. Patients who required assisted ventilation within a given time windows are labelled as "Evol", while those who did not needed NIV are labelled as "No Evol" in this study.
The number "No Evol" patients decreases with the time width. This is justifiable by the fast decline nature of ALS. Most "Evol" patients require NIV within either the first 3 months or between the first 6 and 12 months.

\vspace{-4mm}
\begin{table}[H]
  \caption{Details on ALS dataset for time windows of 90 to 365 days. Class imbalance (per time window) is shown as \% within brackets.}
  \label{tab:freq}
  \begin{tabular}{lcc}
    \toprule
   \textbf{}                & \textbf{Evol (E=1)} & \textbf{No Evol (E=0)} \\
    \midrule
   up to 90 days   & 594 (18\%)               & 2750 (82\%)              \\
90 to 180 days  & 373 (15\%)               & 2186 (85\%)              \\
180 to 365 days & 469 (24\%)               & 1456 (76\%)              \\
  \bottomrule
\end{tabular}
\end{table}

\vspace{-2mm}
From all the tested classifiers, the best results for the prognostic models using a mixture of experts was achieved with the SVM Poly. Therefore, and due to space constraints, we report the results using this classifier. 

\subsection{Learning the base models}

Tables \ref{tab:resultStandard} and \ref{resultsInner} show the classification performance of the base models ($M_{90d}$, $M_{90-180d}$, and $M_{180-365d}$) built with the standard SVM Poly and with Conformal predictors (CPs) coupled with the SVM Poly, respectively. Both the standard classifier and CPs (with $\tau$=0) struggle to accurately identify patients who need NIV, as shown by the low sensitivity values obtained across all time windows. This was already the trickiest class to predict in previous works \cite{Carreiro2015, Pires2018}. Notwithstanding, the sensitivity greatly improves when considering predictions made at high-reliability levels (in this case, predictions with high-credibility values), at the cost of giving a prognostic for a limited number of patients. Using CCPs proved to be useful when predicting NIV in ALS patients, with promising AUC, sensitivity, and specificity values, for high credibility-values, mainly when predicting short ($M_{90d}$) and long-term ($M_{180-365d}$) progressions to respiratory failure.

\subsection{Prognostic model using a mixture of experts}

Table \ref{OverallResults} reports the predictive performance of the proposed prognostic model using a mixture of experts, per reliability-threshold. We aim at simulating a real-world situation, at which, given a new ALS patient, we predict whether he (or she) will need NIV, and the most likely time window of occurrence. The fact of patients not being exclusive of one time windows (i.e., an ALS patient may be "No Evol" in a shorter time windows and become "Evol" on a broader time window), hinders the computation of confusion matrix-based evaluation metrics (such as sensitivity or specificity) since the gold standard is not unique. Nevertheless, we evaluate the number of 1) patients correctly identified as needing NIV in the corresponding time window (labelled as "Evol as Evol"), 2) patients correctly identified as not requiring NIV in the corresponding time window (labelled as "noEvol as noEvol"), and 3) the number of misclassifications due to either a wrong prediction or in an incorrect time window.

Comparing with Table \ref{resultsInner}, we noticed a significant increase in the number of predictions made by the prognostic model using a mixture of experts. This suggests that base models complement each other, and patients who may be hard to classify in one model (and time window) are more similar to the training examples of other models (and time window). The percentage of correct predictions is near 80\% across all time windows. While this performance must be enhanced to be translated to clinical settings, we recall that this model is handling a hard classification task, similar to a multi-task problem. 

\vspace{-2.5mm}
\begin{table}[H]
  \caption{Classification performance of the base models trained with SVM with polynomial kernel within a randomized 5-fold CV scheme, per time windows.}
  \label{tab:resultStandard}
  \begin{tabular}{lccc}
   \toprule
                          & \textbf{AUC} & \textbf{Sensitivity} & \textbf{Specificity} \\* \midrule

\textit{$M_{90d}$}             & 0.794$\pm$0.02   & 0.682$\pm$0.09           & 0.767$\pm$0.02           \\
\textit{$M_{90-180d}$} & 0.684$\pm$0.02   & 0.560$\pm$0.04           & 0.735$\pm$0.02           \\
\textit{$M_{180-365d}$} & 0.752$\pm$0.03   & 0.657$\pm$0.05           & 0.712$\pm$0.03           \\* \bottomrule
\end{tabular}
\end{table}

\vspace{-1.5mm}
\begin{table}
\caption{Classification performance of the base models trained with Cross Conformal Predictors coupled with SVM with polynomial kernel within a randomized 5-fold CV scheme, per time windows and reliability threshold.}
\label{resultsInner}
\begin{tabular}{cccc}
\toprule
\textbf{$M_{90d}$} & \textbf{AUC} & \textbf{Sensitivity} & \textbf{Specificity} \\ \midrule
\textit{All} & 0.801$\pm$0.02 & 0.665$\pm$0.06 & 0.788$\pm$0.02 \\
\textit{0.80} & 0.878$\pm$0.03 (34\%) & 0.863$\pm$0.07 (34\%) & 0.869$\pm$0.01 (34\%) \\
\textit{0.90} & 0.878$\pm$0.02 (19\%) & 0.901$\pm$0.04 (19\%) & 0.845$\pm$0.01 (19\%) \\
\textit{0.95} & 0.899$\pm$0.03 (9\%) & 0.929$\pm$0.04 (9\%) & 0.871$\pm$0.03 (9\%) \\ \midrule
\textbf{$M_{90-180d}$} & \textbf{AUC} & \textbf{Sensitivity} & \textbf{Specificity} \\ \midrule
\textit{All} & 0.706$\pm$0.004 & 0.587$\pm$0.07 & 0.751$\pm$0.02 \\
\textit{0.80} & 0.806$\pm$0.05 (29\%) & 0.759$\pm$0.15 (29\%) & 0.833$\pm$0.03 (29\%) \\
\textit{0.90} & 0.818$\pm$0.04 (18\%) & 0.846$\pm$0.07 (18\%) & 0.808$\pm$0.07 (18\%) \\
\textit{0.95} & 0.842$\pm$0.07 (9\%) & 0.869$\pm$0.09 (9\%) & 0.794$\pm$0.04 (9\%) \\ \midrule
\textbf{$M_{180-365d}$} & \textbf{AUC} & \textbf{Sensitivity} & \textbf{Specificity} \\ \midrule
\textit{All} & 0.766$\pm$0.04 & 0.669$\pm$0.05 & 0.741$\pm$0.03 \\
\textit{0.80} & 0.882$\pm$0.04 (33\%) & 0.842$\pm$0.06 (33\%) & 0.867$\pm$0.03 (33\%) \\
\textit{0.90} & 0.875$\pm$0.03 (20\%) & 0.973$\pm$0.04 (20\%) & 0.847$\pm$0.04 (20\%) \\
\textit{0.95} & 0.886$\pm$0.03 (9\%) & 1.0$\pm$0.0 (9\%) & 0.836$\pm$0.03 (9\%) \\ \bottomrule
\end{tabular}
\end{table}

\vspace{-1.5mm}
\begin{table}[H]
\caption{Classification performance of the prognostic model using mixture of experts trained with Cross Conformal Predictors coupled with SVM with polynomial kernel within a randomized 5-fold CV scheme, per reliability threshold.}
\label{OverallResults}
\begin{tabular}{@{}ccccc@{}}
\toprule
\textbf{} & \textbf{Evol} & \textbf{noEvol} & \textbf{No. (\%) of} & \textbf{\% of} \\
\textbf{$\tau$} & \textbf{as Evol} & \textbf{as noEvol} & \textbf{misclassifications} & \textbf{predictions} \\\midrule
\textit{All} & 561 & 5478 & 1789 (23\%) & 100\% \\
\textit{0.80} & 561 & 5478 & 1789 (23\%) & 100\% \\
\textit{0.90} & 542 & 5429 & 1714 (22\%) & 98\% \\
\textit{0.95} & 410 & 4892 & 1266 (19\%) & 84\% \\ \bottomrule
\end{tabular}
\end{table}

\section{Conclusions}
Early Administration of NIV in ALS patients leads to better prognosis and extended survival times. In this context, we propose a prognostic model to predicts whether, and when, ALS patients would need assisted ventilation, at a given reliability threshold. Such models can be useful to support clinical decisions. High credible predictions can reinforce the decision of prescribing, or not, NIV, and eventually select patients for clinical trials. 

\begin{acks}
This work was supported by FCT through funding of Neuroclinomics2 (PTDC/EEI-SII/1937/2014) and Predict (PTDC/CCI-CIF/ 29877/2017)
projects, research grant (SFRH/ BD/95846/2013) to TP and LASIGE Research Unit (UID/CEC/00408/2019).
\end{acks}

\bibliographystyle{ACM-Reference-Format}
\bibliography{sample-base}

\end{document}